\documentclass[10pt,twocolumn,letterpaper]{article}

\newif\ifarxiv
\arxivtrue  %

\ifarxiv
\usepackage[pagenumbers]{iccv}
\usepackage[absolute,overlay]{textpos}
\else
\usepackage{iccv}
\fi
\usepackage{algorithm}
\usepackage{algorithmic}

\usepackage{adjustbox}
\usepackage{bbold}

\usepackage[scaled=.7]{beramono} %

\definecolor{customblue}{rgb}{0.21,0.49,0.74}
\ifarxiv
\usepackage[breaklinks,colorlinks,citecolor=customblue,bookmarks=true,bookmarksnumbered=true]{hyperref}
\hypersetup{
  pdftitle={SIS-Challenge: Event-based Spatio-temporal Instance Segmentation Challenge 2025},
  pdfsubject={Computer Vision, Robotics},
  pdfauthor={Hamann et al},
  pdfkeywords={Video Instance Segmentation, Event Cameras, Asynchronous Sensor, High Dynamic Range, High Temporal Resolution, Event-based Vision}
}
\else
\usepackage[breaklinks,colorlinks,allcolors=customblue]{hyperref} %
\fi

\usepackage{xcolor}
\definecolor{light-gray}{gray}{0.6}
\newcommand\gframe[1]{{\color{light-gray}\frame{#1}}}

\newcommand{\bnum}[1]{\bfseries #1}

\title{SIS-Challenge: Event-based Spatio-temporal Instance Segmentation Challenge at the CVPR 2025 Event-based Vision Workshop}

\author{Friedhelm Hamann$^{1}$, Emil Mededovic$^{2}$, Fabian Gülhan$^{2}$, Yuli Wu$^{2}$,\\
Johannes Stegmaier$^{2}$, Jing He$^{3}$, Yiqing Wang$^{3}$, Kexin Zhang$^{3}$,\\
Lingling Li$^{3}$, Licheng Jiao$^{3}$, Mengru Ma$^{3}$, Hongxiang Huang$^{4}$,\\
Yuhao Yan$^{5}$, Hongwei Ren$^{4}$, Xiaopeng Lin$^{4}$, Yulong Huang$^{4}$,\\
Bojun Cheng$^{4}$, Se Hyun Lee$^{6}$, Gyu Sung Ham$^{6}$, Kanghan Oh$^{6}$,\\
Gi Hyun Lim$^{6}$, Boxuan Yang$^{7}$, Bowen Du$^{7}$, and Guillermo Gallego$^{1}$\\
\small $^{1}$~TU Berlin, SCIoI, ECDF,
$^{2}$~RWTH Aachen,
$^{3}$~Xidian University,\\
\small $^{4}$~Hong Kong University of Science and Technology,\\
\small $^{5}$~Sun Yat-sen University,
$^{6}$~Wonkwang University,
$^{7}$~Tongji University.
}

\begin{document}
\maketitle

\ifarxiv
\definecolor{somegray}{gray}{0.5}
\newcommand{\darkgrayed}[1]{\textcolor{somegray}{#1}}
\begin{textblock}{11}(2.5, 0.6)
\begin{center}
\darkgrayed{This paper has been accepted for publication at the\\
IEEE International Conference on Computer Vision (ICCV) Workshops, Honolulu, 2025.
\copyright IEEE}
\end{center}
\end{textblock}
\fi

\begin{abstract}
We present an overview of the Spatio-temporal Instance Segmentation (SIS) challenge held in conjunction with the CVPR 2025 Event-based Vision Workshop. 
The task is to predict accurate pixel-level segmentation masks of defined object classes from spatio-temporally aligned event camera and grayscale camera data.
We provide an overview of the task, dataset, challenge details and results.
Furthermore, we describe the methods used by the top-5 ranking teams in the challenge.
More resources and code of the participants' methods are available here: \url{https://github.com/tub-rip/MouseSIS/blob/main/docs/challenge_results.md}
\end{abstract}
    
\section{Introduction}
\label{sec:intro}

With the rapid evolution of computer vision applications in robotics, autonomous systems, and biological research, the ability to accurately segment and track multiple objects over time has become increasingly important.
Traditional frame-based cameras, while widely adopted, face fundamental limitations when dealing with challenging visual conditions such as high-speed motion, varying illumination, and low-light environments.
These limitations are particularly pronounced in applications requiring real-time performance and high temporal precision, such as tracking tasks applicable to many problems, for example, behavioral analysis in neuroscience research and wildlife monitoring.

Event cameras, also known as Dynamic Vision Sensors (DVS)~\cite{Lichtsteiner08ssc,Finateu20isscc}, offer a compelling alternative to conventional frame-based sensors.
Unlike traditional cameras that capture full images at a fixed rate, event cameras asynchronously detect pixel-level brightness changes, producing a sparse stream of events only when and where changes occur in the scene.
This unique sensing paradigm provides several advantages: microsecond temporal resolution, high dynamic range ($>$120 dB), low power consumption, and minimal motion blur~\cite{Gallego20pami}.
These characteristics make event cameras particularly well-suited for tracking fast-moving objects under challenging lighting conditions, where frame-based approaches often fail.

\def\figmethodwidth{.48\linewidth}
\begin{figure}[t]
   \centering
   {\includegraphics[trim={1.8cm 1cm 2.5cm 1cm},clip,width=\linewidth]{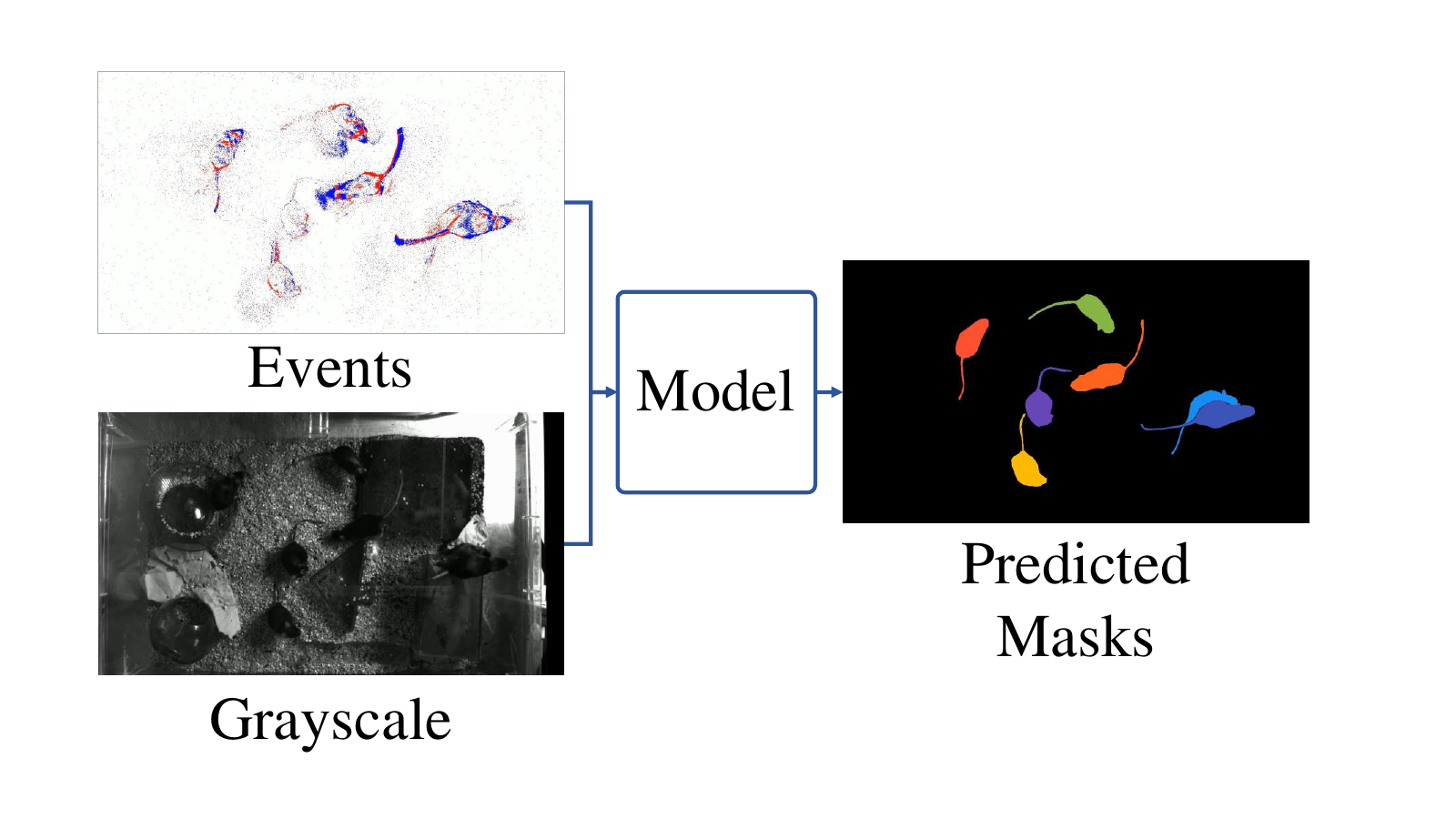}}
\caption{\emph{The Spatio-Temporal Instance Segmentation Challenge}. 
Participants build models that take pixel-level aligned grayscale frames and events as input and predict 
accurate pixel-level segmentation masks and identifiers of all objects of class ``mouse''.}
\label{fig:eyecatch}
\end{figure}

Despite these advantages, the adoption of event-based vision for complex tracking tasks has been limited by the lack of annotated datasets that support fine-grained, multi-object tracking at the pixel level.
Although significant progress has been made in frame-based video instance segmentation~\cite{Voigtlaender19cvpr,Yang19iccv}, the event-based vision community has focused primarily on simpler tasks, such as single-object bounding-box tracking.
This gap in available resources has hindered the development of sophisticated event-based tracking algorithms that could fully leverage the unique properties of event cameras.

To address this critical need, we present the Spatio-temporal Instance Segmentation (SIS) Challenge, organized as part of the 2025 CVPR Event-based Vision Workshop\footnote{\url{https://tub-rip.github.io/eventvision2025/}}.
This challenge is based on the MouseSIS dataset~\cite{Hamann24mousesis}, which provides aligned event and frame data with pixel-accurate instance segmentation masks for multiple freely moving mice.
The dataset includes 33 video sequences with an average duration of approximately 20 seconds, recorded using a beamsplitter system that ensures pixel-level alignment between frames and events.
The sequences contain challenging scenarios, including uneven illumination, occlusions, and complex interactions between multiple targets.

The 2025 SIS Challenge explores algorithmic potentials for multi-object mask-level tracking using event data.
Unlike traditional video instance segmentation tasks that operate on images, the SIS Challenge addresses the unique opportunities presented by the quasi-continuous nature of event streams.
Participants were tasked with developing methods that could accurately segment and track multiple mouse instances throughout entire sequences while maintaining consistent temporal identities (\cref{fig:eyecatch}).
The Challenge ran from February to May 2025, attracting 63 participants with 14 teams submitting results to the leader board.

This paper summarizes the approaches and findings from the top-performing teams in the Challenge.
The results demonstrate that event-based approaches can achieve competitive performance in complex multi-object tracking scenarios, with the winning method achieving a Higher Order Tracking Accuracy (HOTA) score of 0.62.
Hence, the SIS Challenge and this accompanying summary contribute to the advancement of the field of event-based computer vision, showcasing the potential of event cameras for complex scene understanding tasks.
By providing a benchmark for event-based spatio-temporal instance segmentation, we aim to inspire future research in further developing robust tracking algorithms that can operate effectively under challenging visual conditions where traditional cameras struggle.

\section{Spatio-Temporal Instance Segmentation Challenge}
\label{sec:challenge}

\subsection{Introduction of the SIS Dataset}

The SIS Challenge is based on the MouseSIS dataset~\cite{Hamann24mousesis}, a benchmark for multi-object tracking and segmentation using synchronized event and frame data.
The dataset captures freely moving mice in laboratory settings using a specialized hardware setup that ensures pixel-level alignment between neuromorphic event cameras and conventional grayscale cameras.
The MouseSIS dataset comprises 33 sequences, each approximately 20 s in duration: 
around 600 frames at 30 Hz and aligned event data.
The sequences feature varying numbers of mice (1--6 subjects) engaged in natural behaviors under different lighting conditions, including challenging scenarios with occlusions, rapid movements, and uneven illumination.
The dataset follows a YouTubeVIS-style annotation format, providing instance-level segmentation masks with consistent identifiers (IDs) throughout each sequence.
Data is organized into predefined train, validation, and test splits, with sequences distributed to ensure balanced difficulty across splits.

\subsection{Task Description}

The Challenge requires participants to develop algorithms for spatio-temporal instance segmentation of mice from synchronized event and frame data (\cref{fig:eyecatch}). 
Specifically:
\begin{enumerate}
    \item \textbf{Input}: Participants receive pixel-aligned event streams and grayscale frames for each test sequence.
    Event data is provided as raw events $(x, y, t, p)$ where $(x, y)$ are pixel coordinates, $t$ is the timestamp, and $p$ is polarity.
    \item \textbf{Output}: Methods must produce temporally accurate and consistent pixel-level instance segmentation masks 
    and object IDs for all mice in each sequence.
    \item \textbf{Evaluation}: Following the MouseSIS evaluation protocol, methods are assessed using multiple metrics, including HOTA \cite{Luiten21ijcv}, Multiple Object Tracking Accuracy (MOTA) \cite{Bernardin08jivp}, and IDF1 \cite{Ristani16eccv} scores, which jointly evaluate segmentation quality and temporal consistency.
\end{enumerate}

For the Challenge, participants process six test sequences: 10, 16, 22, 26, 28 \& 32.
Sequences 1 and 7 from the original test set \cite{Hamann24mousesis} were excluded to maintain evaluation integrity.
Submissions consist of JavaScript Object Notation (JSON) files containing predicted segmentation masks in Run-Length Encoding (RLE) format with associated instance IDs and confidence scores.

\subsection{Data Loading and Training Pipeline}

To facilitate participation and ensure reproducibility, the Challenge provides a comprehensive codebase with standardized data loading and training pipelines:

\noindent\textbf{Data Access.}
Participants can download the MouseSIS dataset from the provided Google Drive repository, organized in HDF5 format with separate files for each sequence.
Each HDF5 file contains synchronized frames and events with precise temporal alignment information.

\noindent\textbf{Preprocessing Pipeline.}
The codebase includes utilities:
\begin{enumerate}
    \item Loading and synchronizing event and frame data from HDF5 files.
    \item Converting raw events to various representations (e.g., event frames, voxel grids).
    \item Handling the YouTubeVIS-style annotations with proper sequence-instance mapping.
\end{enumerate}

\noindent\textbf{Baseline Implementation.}
A complete baseline method, \emph{ModelMixSort}~\cite{Hamann24mousesis}, that combines YOLOv8 object detection and Segment Anything Model (SAM)-based segmentation with XMem-based tracking, is provided. 
It demonstrates:
\begin{enumerate}
    \item Multi-modal fusion of events and frames.
    \item Integration with popular deep learning frameworks (e.g., PyTorch).
    \item Standard training procedures with configurable hyperparameters.
    \item Inference scripts for generating Challenge-compliant JSON outputs.
\end{enumerate}

The pipeline supports flexible experimentation while maintaining standardized evaluation procedures, enabling a fair comparison of different approaches.

\section{Challenge Results}
\label{sec:results}

This section summarizes the results of the top-5 teams in the Challenge ranking, 
showing an increase of up to 42\% compared to the baseline method \emph{ModelMixSort}~\cite{Hamann24mousesis}.
Most teams follow a similar tracking-by-detection approach as \emph{ModelMixSort}, improving this modular method by integrating and fine-tuning the latest detection and segmentation methods in the literature.
Technical details of all methods can be found in \cref{sec:team_methods}.
In summary, the centralized evaluation and modular baseline method provide easy access, also for non-event-vision practitioners, to the topic of event-based tracking.
This low entry barrier allows participants to integrate the latest advances in foundation models and explore the advantages of event-based cameras.

\begin{table}[ht]
\centering
\begin{adjustbox}{max width=\linewidth}
\begin{tabular}{lccc}
\toprule
\textbf{Team Name} & \textbf{HOTA}$\uparrow$ & \textbf{MOTA}$\uparrow$ & \textbf{IDF1}$\uparrow$ \\
\midrule
1. emilmed & \bnum{0.62} & \bnum{0.72} & \bnum{0.83} \\
2. enidx & 0.57 & 0.69 & 0.75 \\
3. mysterypeople & 0.54 & 0.59 & 0.67 \\
4. shlee & 0.54 & 0.61 & 0.68 \\
5. vivien & 0.54 & 0.54 & 0.67 \\
ModelMixSort (baseline)~\cite{Hamann24mousesis} & 0.43 & 0.45 & 0.50 \\
\bottomrule
\end{tabular}
\end{adjustbox}
\caption{\label{tab:team_results}Top-5 results of the CVPR 2025 Spatio-temporal Instance Segmentation (SIS) Challenge at the Event-based Vision Workshop.
The HOTA score determines the overall ranking.
Bold values indicate the best results per metric.
}
\vspace{-2ex}
\end{table}

\section{Conclusion}
\label{sec:conclusion}

We presented the results of the SIS Challenge held in conjunction with the CVPR'25 Event-based Vision Workshop.
Progress in event-based vision in general, and more specifically in event-based tracking, is lagging behind conventional vision in terms of easily accessible evaluation platforms.
The MouseSIS dataset and the SIS Challenge provide steps towards closing such a gap.
This report provides an overview of the MouseSIS dataset, the challenge, and technical details of the top-5 methods.
The solutions of the participants show creative integration of existing frame-based methods and optimizations, which improve accuracy by $\approx 42\%$ compared to the baseline method.
We believe that this benchmark, which includes an accessible baseline method and centralized evaluation, significantly lowers the entry barrier to event-based tracking and helps foster future developments in this topic.

\ifarxiv
    \section{Challenge Teams and Methods}
    \label{sec:team_methods}
    \subsection{Team 1: emilmed}

\subsubsection{Description}
The tracking pipeline of the winning team (\cref{fig:overview}) draws inspiration from \cite{Seidenschwarz23cvpr}, which emphasizes that carefully selected design choices within a classical tracking-by-detection paradigm can achieve competitive performance, and also highlights the critical role of domain adaptation.

Pretrained Convolutional Neural Networks (CNNs) \cite{Oshea15arxiv}, like those in YOLOv8, rely on filters that assume a certain dynamic range in input intensities. Low-contrast inputs violate this assumption, resulting in weak activations and poorly separated feature maps. 
Histogram equalization improves contrast and reduces the distributional mismatch with COCO-pretrained detectors \cite{Jocher23yolov8}, enhancing the responsiveness of early convolutional filters. 
This constitutes a lightweight form of domain adaptation that aligns low-level input statistics. 
They retrain the detection model on these equalized images and finetune SAM-Large \cite{Kirillov23iccv} using Low-Rank Adaptation (LoRA) \cite{Hu22iclr}, enabling efficient and scalable domain-specific adaptation.

\begin{figure*}[t]
    \centering
    \includegraphics[width=.9\linewidth]{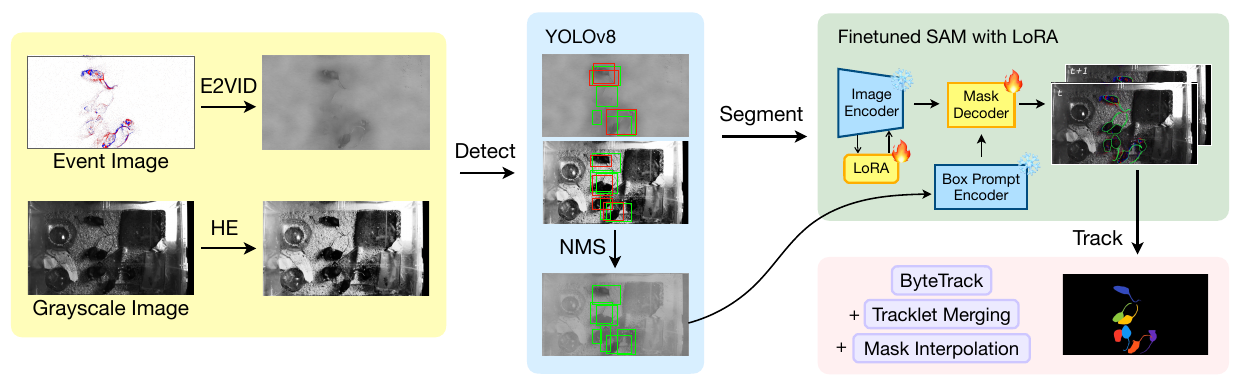}
    \caption{\emph{Team 1}. 
    Overview of the complete tracking pipeline. First, events are reconstructed using E2VID \cite{Rebecq19pami}, while grayscale images are histogram-equalized (HE). 
    These inputs are then passed to individually trained YOLOv8 detectors \cite{Jocher23yolov8}. 
    To eliminate duplicate detections, non-maximum suppression (NMS) is applied based on Intersection over Union (IoU) and detection confidence scores. The resulting bounding boxes are used to prompt a finetuned SAM model \cite{Kirillov23iccv}. 
    Depending on whether the detection originated from the reconstructed or histogram-equalized image, the corresponding input is fed into the SAM encoder for mask prediction. 
    Tracking is performed using the ByteTrack paradigm \cite{Zhang22eccv}, relying solely on bounding boxes. 
    Finally, the tracking output is refined through a greedy merging of fragmented tracks, and missing masks are interpolated to produce the final results.}
    \label{fig:overview}
\end{figure*}

For object association, the team adopts a more classical strategy based on bounding boxes. This approach offers faster and more stable motion prediction compared to mask-level tracking, which can be noisy and computationally demanding in complex scenes. In the following, the key components of the tracking pipeline are elaborated. \\
\textbf{Object Detectors.}
For frame-based input, the team applies histogram equalization during training to mitigate lighting inconsistencies and retrains the YOLOv8 \cite{Jocher23yolov8} detector model using the equalized frames. For event-based input, they use the provided baseline detection model applied to E2VID-reconstructed frames \cite{Rebecq19pami}. \\
\textbf{Segmentation.}
The team finetunes SAM \cite{Kirillov23iccv} by varying only the decoder, leaving the image encoder untouched (i.e., frozen). 
To enable efficient adaptation, they inject LoRA \cite{Hu22iclr} into the encoder's attention projections, modifying weight matrices as follows:
\[
\vspace{-0.5ex}
W \leftarrow W + AB, \quad A \in \mathbb{R}^{d \times r},\ B \in \mathbb{R}^{r \times d},\ r \ll d,
\vspace{-0.5ex}
\]
where $A$, $B$ are trainable and initialized using the method proposed by He et al. \cite{He15iccv}. 
This reduces the trainable parameters while enabling effective task-specific adaptation. \\
\textbf{Post-Processing.}
The team refines the predicted segmentation masks using morphological operations inspired by the SAM2 pipeline \cite{Ravi25iclr}.
Specifically, they apply dilation with a $3 \times 3$ rectangular kernel for 4 iterations, followed by erosion for 3 iterations with the same structuring element.
This operation smooths mask boundaries and fills small internal holes, resulting in more coherent and visually consistent segmentations across frames. \\
\textbf{Tracker.} The method employs ByteTrack \cite{Zhang22eccv_bytetrack} for multi-object tracking and performs data association using a cost matrix that combines Intersection-over-Union (IoU) similarity with detection confidence scores. 
Specifically, it weighs IoU similarities by detection confidence before converting them into a fused cost matrix. 
Appearance-based features were omitted due to the high visual similarity between rodents, which renders embedding-based metrics unreliable.\\
\textbf{Track Merging.} The greedy tracklet merging algorithm used in the pipeline of \cref{fig:overview} is based on temporal proximity and spatial overlap (see Algorithm \ref{alg:track_merge}). 
The procedure considers pairs with small frame gaps and sufficient IoU between their bounding boxes. 
Pairs are greedily merged according to descending IoU.\\
\noindent \textbf{Track Interpolation.} To handle missing detections within a track, the method performs linear interpolation of object centroids at time point $t$ and spatially shifts the segmentation masks. 
Let $M_1 : \mathbb{Z}^2 \to \{0, 1\}$ be a binary mask defined on a 2D pixel grid. 
If there is a gap in the object's tracked trajectory, it is spatially interpolated by translating its centroid.
The team calculates the centroids of the detections at the gap borders using image moments \cite{Gonzalez09book}:
\[
\begin{aligned}
(c_x^{(\tau)}, c_y^{(\tau)}) = ({M_{10}^{(\tau)}}/{M_{00}^{(\tau)}}, {M_{01}^{(\tau)}}/{M_{00}^{(\tau)}}),
\end{aligned}
\]
for $\tau=\{t_1, t_2\}$. 
The interpolated centroid is:
\[
\begin{aligned}
c_x^{\text{interp}} &= (1 - \alpha) c_x^{(t_1)} + \alpha c_x^{(t_2)} \\
c_y^{\text{interp}} &= (1 - \alpha) c_y^{(t_1)} + \alpha c_y^{(t_2)}
\end{aligned}
\quad \text{with} \quad \alpha = \frac{t - t_1}{t_2 - t_1}.
\]

\noindent
Then the required translation is computed:
\[
\Delta x = c_x^{\text{interp}} - c_x^{(t_1)}, \quad
\Delta y = c_y^{\text{interp}} - c_y^{(t_1)}.
\]

The method constructs a 2D affine transformation matrix to perform this translation:
\[
T = \begin{bmatrix}
1 & 0 & \Delta x \\
0 & 1 & \Delta y \\
0 & 0 & 1
\end{bmatrix}
\]
This matrix is applied to $M_{t_1}$ via image warping, yielding the interpolated mask $M_t$ at time $t$, aligned with the linearly estimated object trajectory between frames $t_1$ and $t_2$:
\[
M_t(x, y) = M_1\big(T^{-1} [x,y,1]^\top \big)
\]
This procedure is repeated for each gap frame to reconstruct intermediate segmentation masks.

\begin{algorithm}[t]
\caption{(\emph{Team 1}). Greedy Tracklet Merging Based on Temporal and Spatial Continuity}
\label{alg:track_merge}
\begin{algorithmic}[1]
    \REQUIRE Tracklets $\mathcal{T} = \{T_i\}$ with $(s_i, e_i, B^{\text{first}}_i, B^{\text{last}}_i)$ 
    \ENSURE Merged tracks
    \STATE Initialize empty list of merge candidates $\mathcal{C}$
    \FOR{all pairs $(T_i, T_j)$ with $i \neq j$}
        \STATE Compute frame gap $\Delta_{ij} = s_j - e_i$
        \IF{$\Delta_{\min} < \Delta_{ij} \leq \Delta_{\max}$}
            \STATE Compute $\text{IoU}_{ij} = \text{IoU}(B^{\text{last}}_i, B^{\text{first}}_j)$
            \IF{$\text{IoU}_{ij} \geq \theta$}
                \STATE Add $(i, j)$ to candidate list $\mathcal{C}$
            \ENDIF
        \ENDIF
    \ENDFOR
    \STATE Sort $\mathcal{C}$ by descending IoU
    \FOR{each pair $(i, j) \in \mathcal{C}$}
        \IF{$T_i$ and $T_j$ are not already merged}
            \STATE Merge $T_i$ and $T_j$ into a single track
        \ENDIF
    \ENDFOR
\end{algorithmic}
\end{algorithm} 

\subsubsection{Implementation Details}
\textbf{Object Detector.} The team retrained the YOLOv8m object detector \cite{Jocher23yolov8} on the provided MouseSIS dataset \cite{Hamann24mousesis}. 
The model was initialized with pretrained weights and trained for 100 epochs with a batch size of 16. Input images were resized to 640 pixels on the longest side while preserving aspect ratio. Optimization was performed using an SGD optimizer (with momentum 0.937 and weight decay 0.0005), with an initial learning rate of 0.01 and a 3-epoch warm-up phase, as per the default setting. To improve generalization, they applied several augmentations during training, including Mosaic augmentation, horizontal flipping with a probability of 0.5, random translations (±10\%), scaling (±50\%) and random erasing with a probability of 0.4. \\
\textbf{Segmentation.} The team finetuned SAM \cite{Kirillov23iccv} using the MouseSIS dataset \cite{Hamann24mousesis}. 
The model was initialized with the official \texttt{sam\_vit\_l\_0b3195} checkpoint. 
LoRA modules with rank \texttt{\( r = 16 \)} were added to the query-key-value projections of selected self-attention layers in the image encoder. 
During training, only the LoRA modules and the mask decoder were updated, while the rest of the model remained frozen. 
Each image was resized to 1024 px on the longest side, normalized to \([0, 1]\), and padded to 1024$\times$1024 px. 
Training was carried out using the Adam optimizer with a learning rate of \(10^{-5}\) for up to 500 epochs, with gradient accumulation every 8 steps to simulate a larger batch size. 
The loss combined binary cross-entropy and Dice loss (weighted 0.005). 
The team manually applied early stopping based on validation loss. \\
\textbf{Tracking.} During tracking, the team performed non-maximum suppression between bounding boxes predicted from the E2VID reconstructions and frames. 
The finetuned SAM model was then applied to the E2VID input, as it better matches the training domain compared to the off-the-shelf model. 
They adjusted the tracker's association parameters to suit the task, as follows:
\texttt{track\_high\_thresh = 0.6}, \texttt{track\_low\_thresh = 0.1}, \texttt{match\_thresh = 0.8}, \texttt{new\_track\_thresh = 0.7}, \texttt{track\_buffer = 60}. 
They set the temporal association bounds to \texttt{$\Delta_{\min} = -15$} and \texttt{$\Delta_{\max} = 15$}, and used an IoU threshold \texttt{$\theta = 0.1$} for linking tracklets.

All training and inference runs have been performed on a single NVIDIA RTX 3090 graphics card. 

\subsubsection{Results}
The proposed pipeline achieved first place on the MouseSIS challenge. 
The results are reported in \cref{tab:team_results}. 
The team outperformed other proposed solutions in all tracking metrics.
However, the computation time is relatively high (2 seconds per sample, excluding E2VID runtime, as images are reconstructed in advance) and could benefit from model distillation to improve efficiency.

    \subsection{Team 2: enidx}

\subsubsection{Description}

The team proposes a series of enhancements based on ModelMixSort, a tracking-by-detection approach in~\cite{Bewley16icip}, to improve tracking and segmentation performance. 
Specifically, they replace YOLOv8 with the more powerful YOLOv12 for object detection, and substitute SAM2 with an upgraded version of SAM to enhance feature extraction capabilities. 
For preprocessing, they apply contrast enhancement to grayscale frames, and during inference, they employ test-time augmentations such as image rotation and flipping to improve model robustness and generalization. 
As a result, their method achieves great performance on the test set, ranking second overall (\cref{tab:team_results}): HOTA$=$0.569, MOTA$=$0.688 and IDF1$=$0.748. 

\subsubsection{Implementation Details}

\paragraph{Input Process.}

The MouseSIS dataset captures the activities of multiple mice in complex environments using synchronized event and frame cameras~\cite{Hamann24mousesis}.
It provides high-quality instance segmentation masks, bounding boxes, and identity annotations for each mouse across video frames.
The dataset includes numerous challenging scenarios involving frequent occlusions and interactions, making it well-suited for evaluating robust multi-object tracking and segmentation methods.

To improve the performance of downstream object detection and instance segmentation models, the team applies contrast enhancement as a standardized preprocessing step for all input grayscale frames.
This design addresses the issue of weak feature visibility and blurred object boundaries, which are common in low-light or low-contrast scenes, and often degrade the discriminative capability of deep-learning models.
Specifically, they adopt the Contrast Limited Adaptive Histogram Equalization (CLAHE) method with parameters clipLimit 2.0 and tileGridSize (8,8) to enhance the local contrast of each frame.
By redistributing the grayscale values within localized regions, CLAHE expands the dynamic range and improves the visibility of fine details and textures that are otherwise hard to detect. 

Contrast enhancement is applied consistently across the training, validation, and testing sets to ensure feature distribution alignment throughout the entire pipeline.
This consistency helps the model learn stable representations and mitigates performance degradation due to distribution shift.

\noindent \textbf{Boxes Detection.}
To achieve accurate detection of mice, the team adopts YOLOv12 \cite{Tian25yolov12}, a recent advancement over YOLOv8 with improved detection performance and feature extraction capability.
Specifically, they use the large-scale YOLOv12-X variant to take advantage of its enhanced representational power.
Given the multimodal nature of the data, they train two separate YOLOv12-X detectors: one for the event data and another for the grayscale frames. 
For the event modality, they first reconstruct frames from raw event streams using E2VID \cite{Rebecq19pami}.

The dataset was partitioned by sequence, with frames 400 to 757 from sequence 33 designated as the validation set, while all remaining frames from the validation dataset were used for training. 
During training, all input images were uniformly resized to 640\texttimes640 pixels.
The model was initialized with YOLOv12-X weights pretrained on COCO to accelerate convergence and enhance generalization performance. 
A batch size of 8 was employed during training, and each detector was trained for 80 epochs on its corresponding dataset.
All other training configurations, including the choice of optimizer, learning rate scheduling policy, strictly followed the default settings provided by the official YOLO implementation to ensure consistency and reproducibility. 

\noindent \textbf{Segmentation and Tracking.}
After obtaining high-quality detection boxes from the YOLOv12-X, the team further employed SAM2 \cite{Ravi25iclr} for fine-grained instance segmentation.
Specifically, they selected the more advanced SAM2.1\_hiera\_large, which demonstrates significantly improved segmentation performance compared to the original SAM \cite{Kirillov23iccv}.
The detection boxes produced by YOLOv12-X were used as prompts to guide SAM2 in generating corresponding segmentation masks.
For the video object segmentation model, they used XMem \cite{Cheng22cvpr}, set to follow \cite{Hamann24mousesis}.

In order to make SAM2 better adapt to the specific appearance and pose variations of the mice in the MouseSIS dataset, the team fine-tuned the model according to the officially published pre-training weights. The fine-tuned dataset consists of MouseSIS training and validation datasets of grayscale frames.
 
The fine-tuning was performed using four NVIDIA GeForce RTX 4090 GPUs, with the batch size set to 1, the maximum number of objects processed per image capped at 6, and the resolution of all input images set to 1280 px, taking into account the GPU memory limitations and the model's requirement for multi-object segmentation.
Base learning rate was set to $5\cdot 10^{-6}$, and the model was trained for a total of 40 epochs, with the rest of the training parameters following the default settings recommended in the SAM2 open-source implementation.

\paragraph{Test-Time Augmentation.}
In order to improve the stability and accuracy of the detection frame and thus enhance the segmentation of the SAM2, the team designed and implemented a test-time enhancement strategy. The strategy works by applying geometric transformations to the input images, such as horizontal flip, scaling, and small-angle rotation, and performing YOLOv12-X detection on each transformed image. All detections are mapped back to the original image coordinate system by inverse transformation and fused with the original detection frames to obtain a more robust bounding box, which is used as a segmentation cue input to SAM2.

During the fusion process, they use the Hungarian algorithm to match the detected frames under different transformations by the intersection and concurrency ratio, and only retain the frames with $\text{IoU}>0.3$ and with area changes within a reasonable range.
All matched frames for each target are weighted and averaged according to the confidence level to generate the final bounding box.
The fusion results are fed into the SAM2 to obtain a more accurate instance segmentation mask.
This method improves detection stability and helps improve the performance of metrics, such as HOTA, MOTA, and IDF1, in multi-target segmentation tasks.

\subsubsection{Results}

The team conducted a series of experiments on the MouseSIS dataset to evaluate the impact of different combinations of detection and segmentation models on the performance of multi-target tracking and segmentation (MOTS).

\Cref{tab:main_results_enidx} shows the impact of different detector and segmentation model combinations on MOTS performance metrics.
Replacing the segmentation module from the original SAM to the SAM2 under the YOLOv8 detector yields substantial improvements: HOTA increases by 8.44\%, MOTA by 10.74\%, and IDF1 by 11.98\%.
This indicates that the enhanced segmentation model significantly improves the quality of instance masks, which in turn benefits tracking accuracy.
Building on this, replacing the detector from YOLOv8 to YOLOv12 results in additional gains of 0.69\% in HOTA, 1.69\% in MOTA, and 1.72\% in IDF1, highlighting the importance of more precise and stable detection boxes in supporting segmentation and target association. 

Furthermore, the use of CLAHE-based contrast enhancement, indicated by an asterisk in \cref{tab:main_results_enidx} as YOLOv12*, results in additional improvements of 0.77\% in HOTA, 0.55\% in MOTA, and 1.37\% in IDF1. 
These results demonstrate that progressive enhancements of the segmentation model, detection accuracy, and input quality collectively contribute to better MOTS performance.

\begin{table}[t]
\centering
\begin{adjustbox}{max width=\linewidth}
\begin{tabular}{@{}lllccc@{}}
\toprule
\textbf{Det. Model} & \textbf{Seg. Model} & \textbf{Tracker} & \textbf{HOTA}$\uparrow$ & \textbf{MOTA}$\uparrow$ & \textbf{IDF1}$\uparrow$\\
\midrule
Yolov8   & SAM  & XMem & 0.431 & 0.445 & 0.500\\
Yolov8   & SAM2 & XMem & 0.516 & 0.552 & 0.620\\
Yolov12  & SAM2 & XMem & 0.522 & 0.569 & 0.637\\
Yolov12* & SAM2 & XMem & 0.530 & 0.575 & 0.651\\
\bottomrule
\end{tabular}
\end{adjustbox}
\caption{(\emph{Team 2}) Comparison of MOTS performance of different model combinations on MouseSIS test dataset.\label{tab:main_results_enidx}}
\end{table}

\begin{table}[t]
\centering
\setlength{\tabcolsep}{4pt} %
\begin{adjustbox}{max width=\linewidth}
\begin{tabular}{@{}llccc@{}}
\toprule
\textbf{Det. Model} & \textbf{Seg. Model} & \textbf{HOTA}$\uparrow$ & \textbf{MOTA}$\uparrow$ & \textbf{IDF1}$\uparrow$\\
\midrule
YOLOv12             & SAM2              & 0.535 & 0.575 & 0.650 \\
YOLOv12+Val         & SAM2              & 0.540 & 0.572 & 0.675 \\
YOLOv12+Val         & SAM2 (fine-tuned) & 0.542 & 0.669 & 0.695 \\
YOLOv12+Val+TTA     & SAM2 (fine-tuned) & 0.569 & 0.688 & 0.748 \\
\bottomrule
\end{tabular}
\end{adjustbox}
\caption{(\emph{Team 2}) Impact of Training and Inference Strategies on MOTS Performance. \label{tab:enidx:impact}}
\vspace{-2.5ex}
\end{table}

Based on the use of the YOLOv12 detector, the SAM2 segmentation model, and CLAHE preprocessing, the team further explored the effect of replacing the original XMem tracker with XMem-no-sensory weights (\cref{tab:enidx:impact}).
To ensure the adequacy of training, the configuration labeled ``+val'' in the table indicates that only the data after the 400th frame in the 33rd sequence is used as the validation set, and the rest of the original validation dataset data are used to train YOLOv12.
This segmentation strategy can better utilize the labeled data and effectively improve model performance.

In addition, the team fine-tuned the SAM2 segmentation model on the MouseSIS dataset to make it more adaptive to the specific task scenarios (\cref{tab:enidx:impact}), resulting in an increase of MOTA by 9.65\%, and the introduction of the test-time-enhancement (TTA) strategy for the detection frames continued to optimize the performance of the model, and the HOTA increased to 0.569, the MOTA reached 0.688, and the IDF1 reached 0.748.
The experiment further verifies the effectiveness of the multi-stage optimization strategy in the multi-target segmentation tracking task.

    \subsection{Team 3: mysterypeople}

\subsubsection{Introduction}
The team proposes a hybrid approach combining the event-image fusion segmentation framework EvInsMOS~\cite{Wan25evinsmos} and the tracking-by-detection methodology inspired by the XMem-based pipeline introduced in the MouseSIS benchmark~\cite{Hamann24mousesis}.

\subsubsection{Method Overview}
As shown in \cref{fig:pipeline}, the pipeline integrates the strengths of both segmentation and tracking:
\begin{itemize}
  \item The team employs the EvInsMOS model as the backbone segmentation network, which fuses texture features from grayscale images and motion cues from event voxels.
  \item To reduce missed instance detection and improve spatial localization, they enhance the decoder with an additional bounding box regression head supervised by $L^1$ loss.
  \item For temporal consistency and identity assignment, they apply an XMem-based tracker~\cite{Cheng22cvpr} to the segmentation outputs, enabling cross-frame association.
\end{itemize}

\begin{figure}[t]
  \centering
  \includegraphics[width=\linewidth]{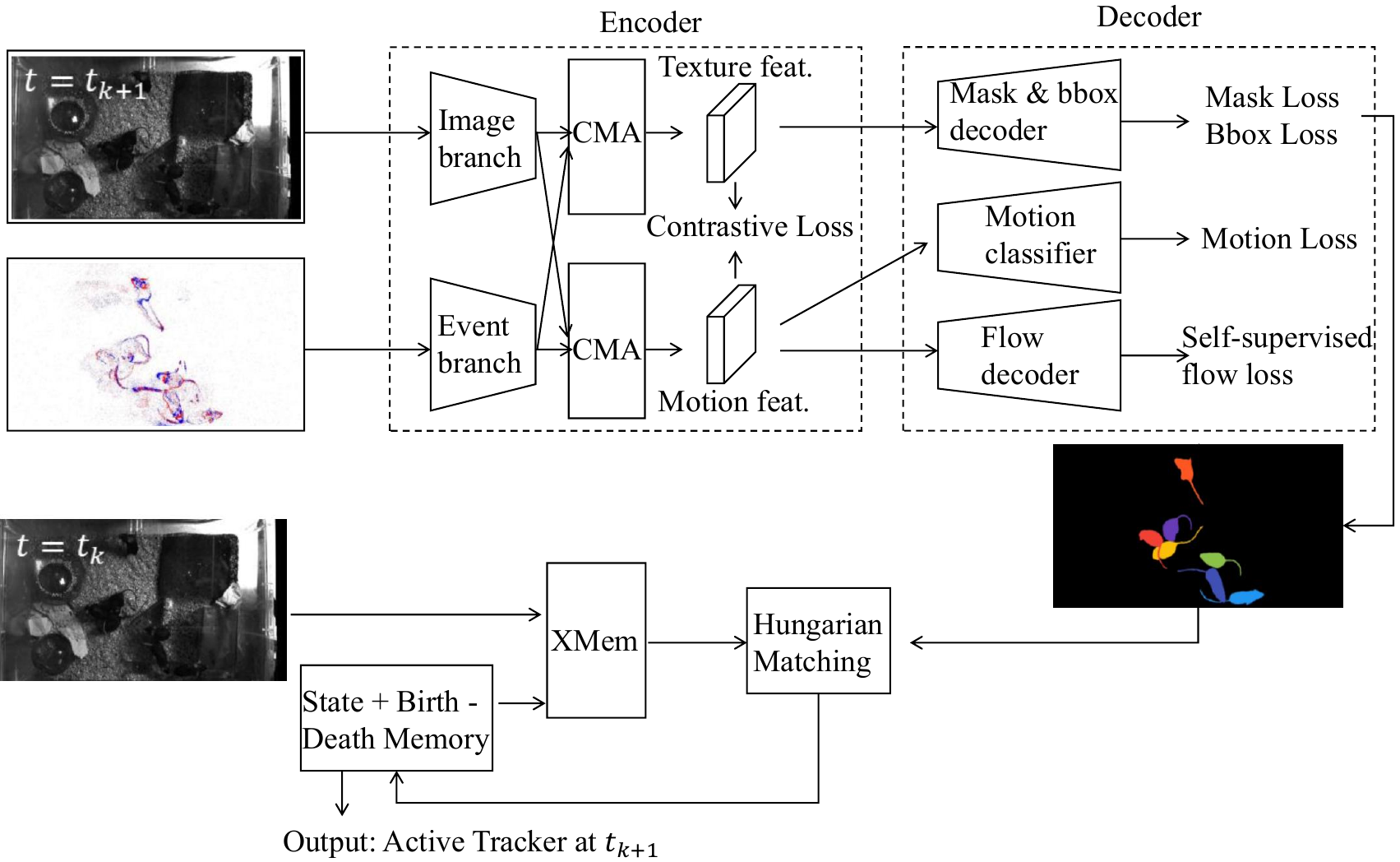} 
  \caption{\emph{Team 3}. Overall pipeline of the method combining EvInsMOS and XMem tracking.}
  \label{fig:pipeline}
\end{figure}

\subsubsection{Segmentation Network Design}

\textbf{Encoder.}
The encoder consists of two modality-specific branches:
\begin{itemize}
  \item \textbf{Image Branch:} A ResNet-based encoder extracts high-resolution texture features $f_I$.
  \item \textbf{Event Branch:} Event streams are voxelized into $H \times W \times B$ tensors (with $B=10$ bins), following Eq.~(1) from~\cite{Wan25evinsmos}. These are processed by a lightweight CNN to obtain motion features $f_E$.
\end{itemize}

These two feature maps are augmented via the \textbf{Cross-Modal Masked Attention (CMA)} module:
\begin{align}
  f_T &= (f_E \odot f_{EM}) \cdot \mathrm{softmax}(f_I^\top (f_E \odot f_{EM}) / \tau) + f_I \\
  f_M &= f_I \cdot \mathrm{softmax}((f_E \odot f_{EM})^\top f_I / \tau) + f_E
\end{align}
where $f_{EM}$ is an event mask, $\tau$ is a learnable temperature, and $\odot$ denotes element-wise multiplication.

\textbf{Contrastive Learning Objective:} The team adopts a multi-frame InfoNCE-based contrastive learning strategy, following the design in EvInsMOS~\cite{Wan25evinsmos}. Specifically:
\begin{itemize}
  \item \textbf{Positive pairs} are constructed between feature representations of the same modality (either $f_T$ or $f_M$) across adjacent frames in a batch.
  \item \textbf{Negative pairs} include features from different modalities (i.e., motion vs. texture) as well as from non-adjacent samples across the batch.
\end{itemize}
The goal is to enforce inter-frame temporal consistency for each modality while encouraging decorrelation between texture and motion modalities. The total contrastive loss is:
\begin{equation}
\mathcal{L}_\text{cl} = - \frac{1}{B} \sum_{b=1}^{B} \log \left(\frac{\mathrm{FC}_T^b + \mathrm{FC}_M^b}{\mathrm{SS}_T^b + \mathrm{SS}_M^b + \mathrm{CS}_{T,M}^b}\right)
\end{equation}

In this formulation, $B$ denotes the number of samples in the mini-batch. The terms are defined as follows:
\begin{itemize}
  \item $\mathrm{FC}_T^b$ and $\mathrm{FC}_M^b$ represent the feature consistency scores between the current and reference frame within the texture and motion modalities, respectively. These scores correspond to positive pairs.
  \item $\mathrm{SS}_T^b$ and $\mathrm{SS}_M^b$ denote the self-similarity scores within the same modality but from unrelated samples, used as intra-modality negative pairs.
  \item $\mathrm{CS}_{T,M}^b$ is the cross-modality similarity score between texture and motion features across the batch, used to penalize shared representations between distinct modalities.
\end{itemize}
This formulation ensures temporal coherence within modalities while enforcing modality-specific representations.

\textbf{Decoder.}
The decoder adopts a query-based design inspired by Mask2Former~\cite{Cheng22masked}, which decouples segmentation and classification into separate parallel branches. The team generates $n$ fixed learnable queries that attend to different object instances.

\begin{itemize}
  \item \textbf{Mask Decoder:} Takes the augmented texture feature $f_T$ and projects each query embedding into a segmentation mask $\hat{S}_i$. Here, $n$ denotes the total number of queries.
  \item \textbf{Motion Classifier:} Takes the augmented motion feature $f_M$ and predicts a binary motion score $\mathbf{m}_i \in \{0,1\}$ for each query, indicating whether the corresponding instance is moving.
\end{itemize}

Decoupling these tasks is beneficial because motion state may not perfectly align with spatial contours, especially in the presence of camera-induced parallax. This separation allows motion classification to benefit from motion-specific cues and segmentation to focus on spatial accuracy.

\begin{table}[ht]
\centering
\begin{adjustbox}{max width=\linewidth}
\begin{tabular}{@{}lccc@{}}
\toprule
\textbf{Methods} & \textbf{HOTA}$\uparrow$ & \textbf{MOTA}$\uparrow$ & \textbf{IDF1}$\uparrow$ \\
\midrule
ModelMixSort (baseline)~\cite{Hamann24mousesis} & 0.43 & 0.45 & 0.50 \\
EvInsMOS + bbox + Hungarian & 0.509 & 0.498 & 0.605 \\
EvInsMOS + XMem tracker & 0.524 & 0.579 & 0.640 \\
EvInsMOS + bbox + XMem (\emph{Team 3}) & \textbf{0.542} & \textbf{0.594} & \textbf{0.669} \\
\bottomrule
\end{tabular}
\end{adjustbox}
\caption{ (\emph{Team 3}) Performance comparison on original resolution (1280\texttimes720 px) MouseSIS test set.}
\label{tab:performance}
\end{table}

\textbf{Training:} For each frame, ground truth instance masks $S_{gt}^{(i)}$ and their motion labels $c_{gt}^{(i)}$ are assigned to predicted queries using Hungarian matching $\rho(i)$ based on spatial IoU. The combined loss is:
\begin{equation}
\mathcal{L}_\text{mos} = \sum_{i=1}^{n} \Bigl( \mathcal{L}_\text{ce}(\mathbf{m}_{\rho(i)}, c_{gt}^{(i)}) + \mathbb{1}_{c_{gt}^{(i)}=1} \cdot \mathcal{L}_\text{mask}(\hat{S}_{\rho(i)}, S_{gt}^{(i)}) \Bigr)
\end{equation}
where $\mathcal{L}_\text{mask}$ is a mixture of focal and dice losses, and $\mathcal{L}_\text{ce}$ is binary cross entropy for motion label prediction.

Also, the team adds a bounding box regression loss:
\begin{equation}
\mathcal{L}_\text{bbox} = \sum_{i=1}^n \left\| \hat{b}_i - b_i^{gt} \right\|_{1}
\end{equation}

\textbf{Inference:} At test time, all $n$ predicted masks $\hat{S}_i$ and motion scores $\mathbf{m}_i$ are computed. Masks with $\mathrm{softmax}(\mathbf{m}_i) > \theta$ are retained as the final $m$ moving instance predictions. This enables the model to adaptively determine the number of foreground instances.

To further enhance the decoder's sensitivity to motion boundaries, the team incorporates an optical flow-guided feature modulation mechanism. 
An unsupervised flow estimator computes the optical flow between adjacent grayscale frames. 
The resulting flow field is used to warp and align decoder-level feature maps across time. 
This flow-guided alignment enhances temporal coherence and sharpens motion contours, particularly in occlusion-prone or fast-motion regions. 
The warped feature $f_\text{warp}$ is fused with the decoder output via attention-based gating, refining the segmentation quality without requiring additional supervision.

\subsubsection{Tracking and ID Association}
The team uses the XMem~\cite{Cheng22cvpr} tracker, specifically they:
\begin{enumerate}
  \item Generate per-frame masks from EvInsMOS.
  \item Feed masks and frames to XMem for propagation.
  \item Use Hungarian matching based on IoU to align new masks with existing memory trackers.
\end{enumerate}

\subsubsection{Implementation Details}
The implementation is in PyTorch 2.5.1, on four NVIDIA A40 GPUs with batch size 8, Adam optimizer, an initial learning rate of $10^{-4}$ trained for 300,000 iterations.
For the XMem-based tracker, the team uses the following hyperparameter configuration: the maximum age of a tracker (`max\_age`) is set to 1,
`min\_hits = 3`
The IoU threshold
is set to 0.5 (`iou\_threshold = 0.5`).

\subsubsection{Results}

The team compares their method against three alternative settings: 
(1) ModelMixSort baseline as proposed in \cite{Hamann24mousesis}; 
(2) EvInsMOS with added bounding box regression and Hungarian matching for association; 
(3) EvInsMOS combined with XMem-based tracker. 
Their full model integrates both the enhanced decoder and XMem tracking.

From \cref{tab:performance}, it can be observed that the method outperforms all baselines across all three metrics.  
The addition of the bounding box regression head (comparing row~3 with row~4) contributes to better localization and motion estimation. 
Replacing Hungarian matching with the memory-based XMem tracker (comparing row~2 with row 4) boosts identity preservation as reflected in IDF1. 
Combining both enhancements results in the best performance.

    \subsection{Team 4: Shlee}

\subsubsection{Description}

To address the challenge \cite{Hamann24mousesis}, the team specifically focuses on handling background noise in existing event-based instance segmentation methods. Event cameras inherently produce various types of noise, but here they focus on background noise, especially that caused by nearby light sources, which degrade performance in tasks such as object detection, tracking, and segmentation \cite{Guo22pami,Lim24ur}. In E2VID \cite{Rebecq19pami}, during voxelization, the number of events in each voxel directly influences the quality of the reconstructed image. To mitigate this issue, the team employs a two‐component Gaussian mixture model \cite{Dempster77rss} to separate noisy events from informative ones, resulting in low- and high-frequency clusters. The filtered events are then converted into reconstructed images via E2VID. These images are passed to an object detector to produce denoised intensity images.

\begin{figure}[tbh]
    \centering
    \begin{tabular}{cc}
        \includegraphics[width=0.45\linewidth, keepaspectratio]{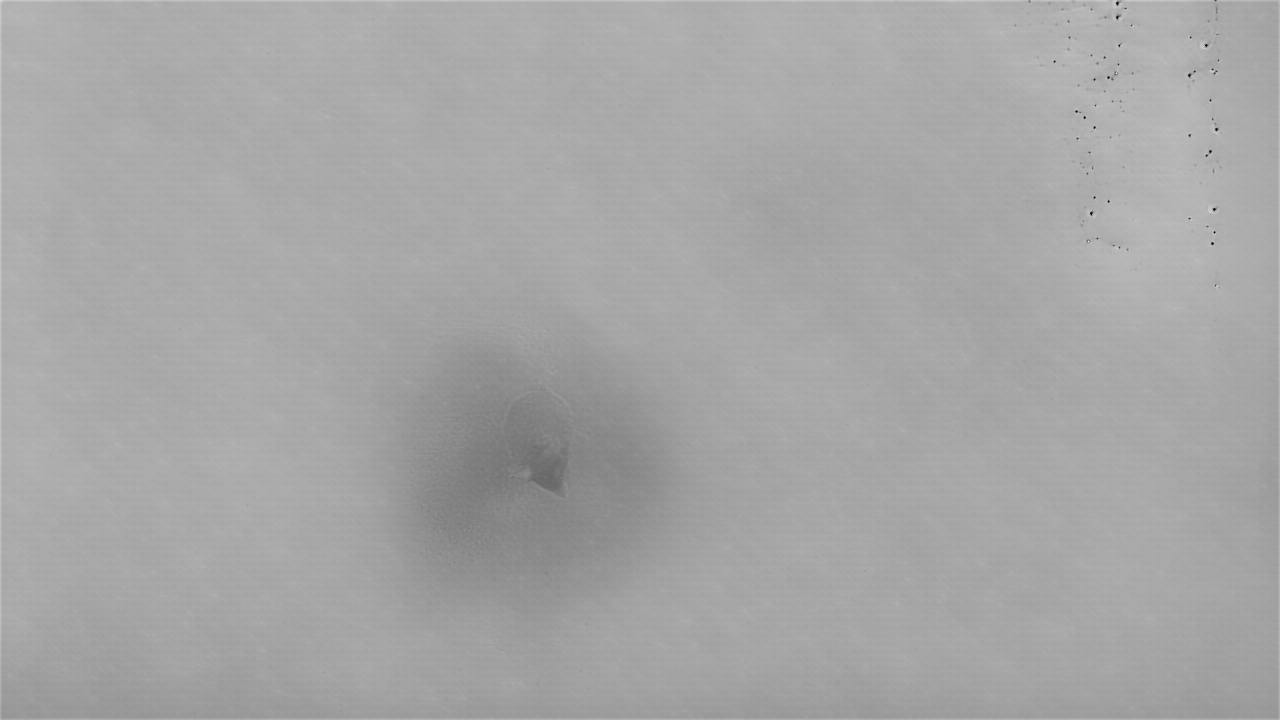} &
        \includegraphics[width=0.45\linewidth, keepaspectratio]{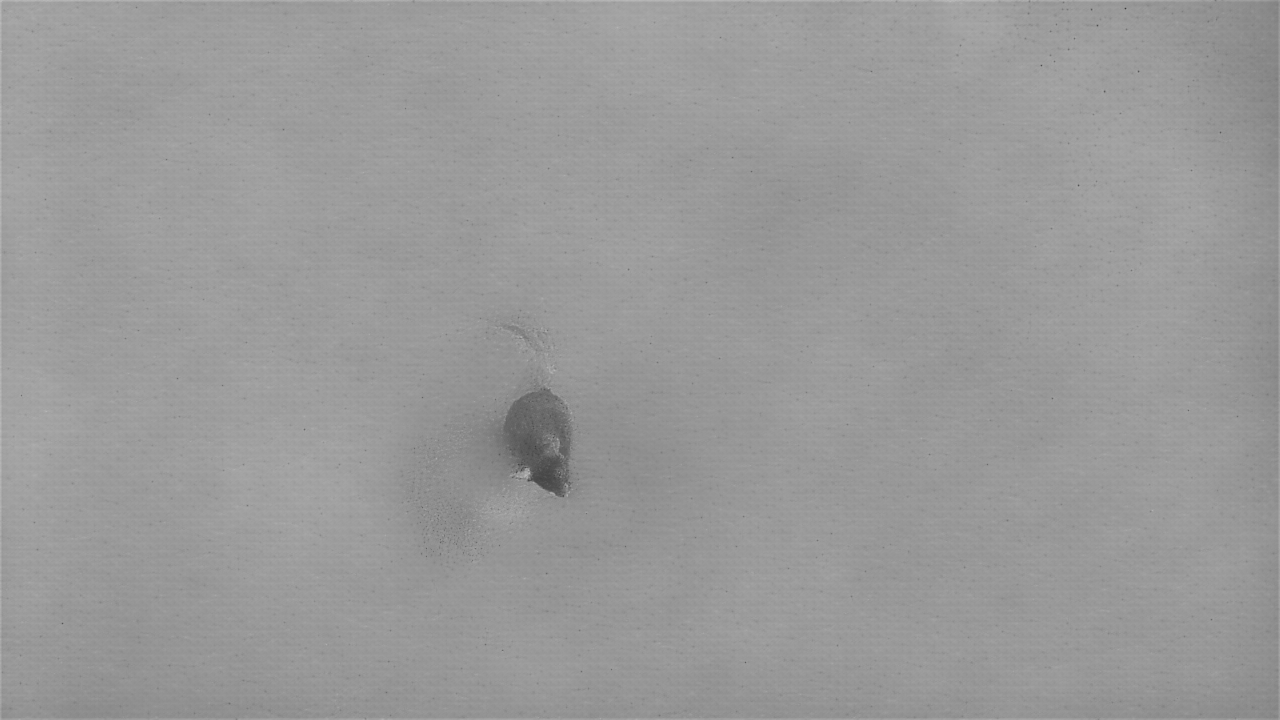} \\
        (a) Baseline Reconstruction &
        (b) Team's Reconstruction
    \end{tabular}
    \caption{(\emph{Team 4}). Comparison between the baseline frame (a) and the reconstructed frame (b).}
    \label{fig:frame_comparison}
\end{figure}

These images are used to fine‐tune a YOLOv8 \cite{Jocher23yolov8} detector, which outputs bounding boxes for each object. The resulting bounding boxes serve as prompts for the SAM2 \cite{Ravi25iclr} model, which predicts instance segmentation masks. Finally, an XMem \cite{Cheng22cvpr} tracker is applied to link object instances across frames, producing a unified spatio‐temporal segmentation and tracking output.

\subsubsection{Implementation Details}
Data processing begins by identifying, for each image frame, the first event whose timestamp matches that frame. 
The team then applies a fixed‐size event count window of 30,000 events centered on this event, collecting events both before and after to form the input set. 
This event subset is then passed to their event‐count‐based GMM clustering (\(k = 2\)), which generates a low‐frequency cluster and a high‐frequency cluster, with means \(\mu_1\) and \(\mu_2\), respectively. 
They compute the absolute difference of these means,
\begin{equation}\label{eq:delta_mu}
\Delta \mu = \left| \mu_1 - \mu_2 \right|,
\end{equation}
and compare it to a threshold \(\tau = 2.5\). 
In other words:
\begin{equation}\label{eq:voxel_rule}
\text{Voxel} =
\begin{cases}
\text{Select a low frequency cluster}, & \Delta \mu \ge \tau,\\
\text{Select both clusters},           & \Delta \mu < \tau.
\end{cases}
\end{equation}
After selecting clusters according to this rule, they voxelize the chosen events and feed them into the E2VID model for reconstruction, as shown in \cref{fig:frame_comparison:two}.

\begin{figure}[t]
    \centering
    \begin{tabular}{cc}
        \gframe{\includegraphics[width=0.45\linewidth, keepaspectratio]{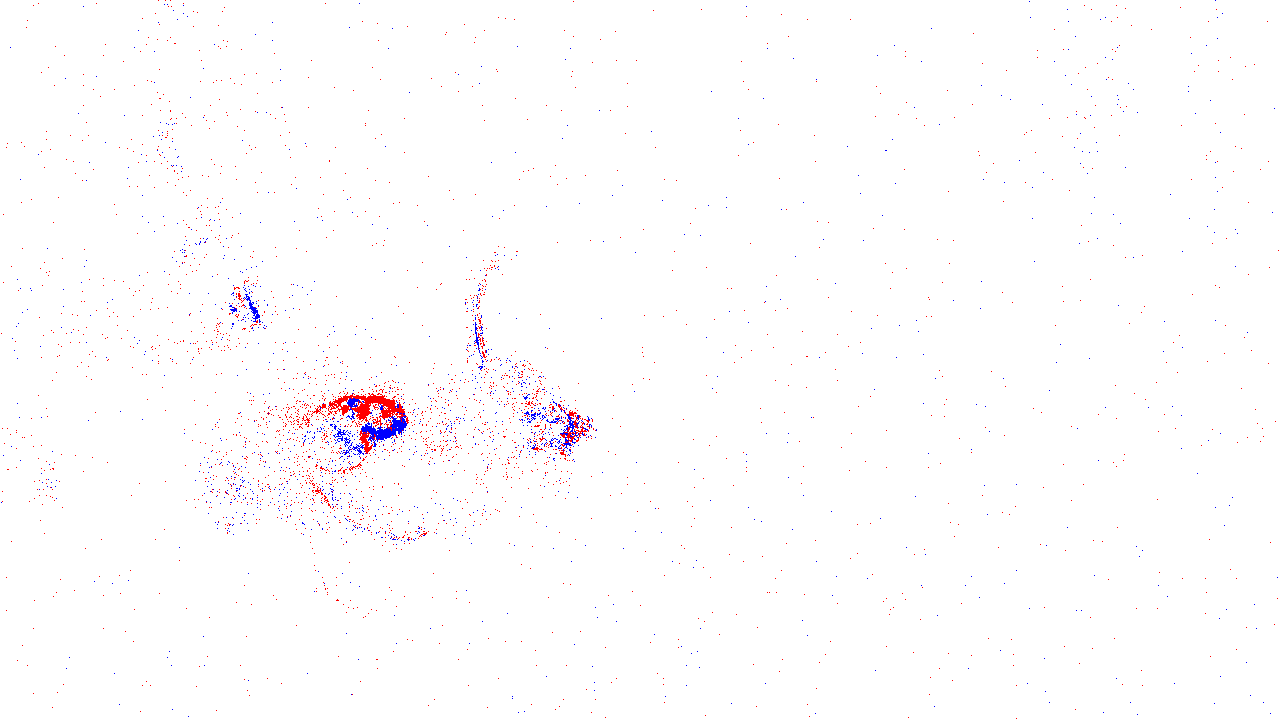}} &
        \gframe{\includegraphics[width=0.45\linewidth, keepaspectratio]{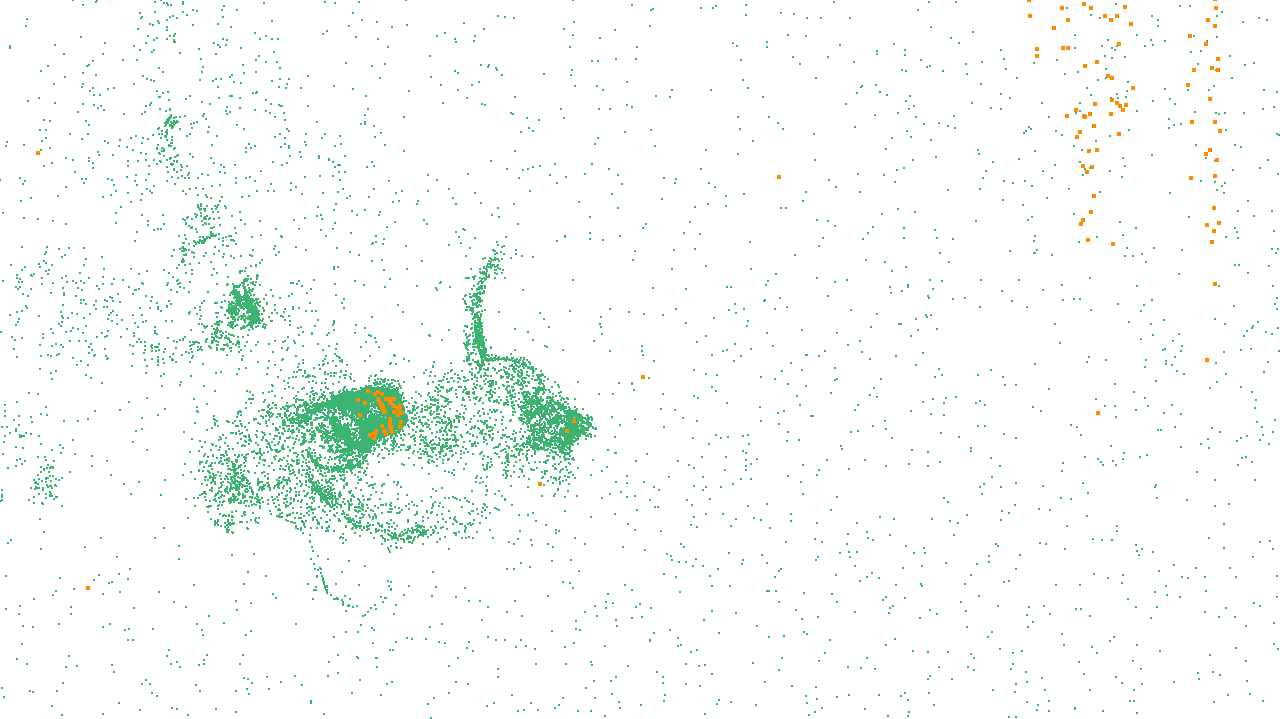}} \\
        (a) Events &
        (b) GMM Cluster \\[6pt]
        \includegraphics[width=0.45\linewidth, keepaspectratio]{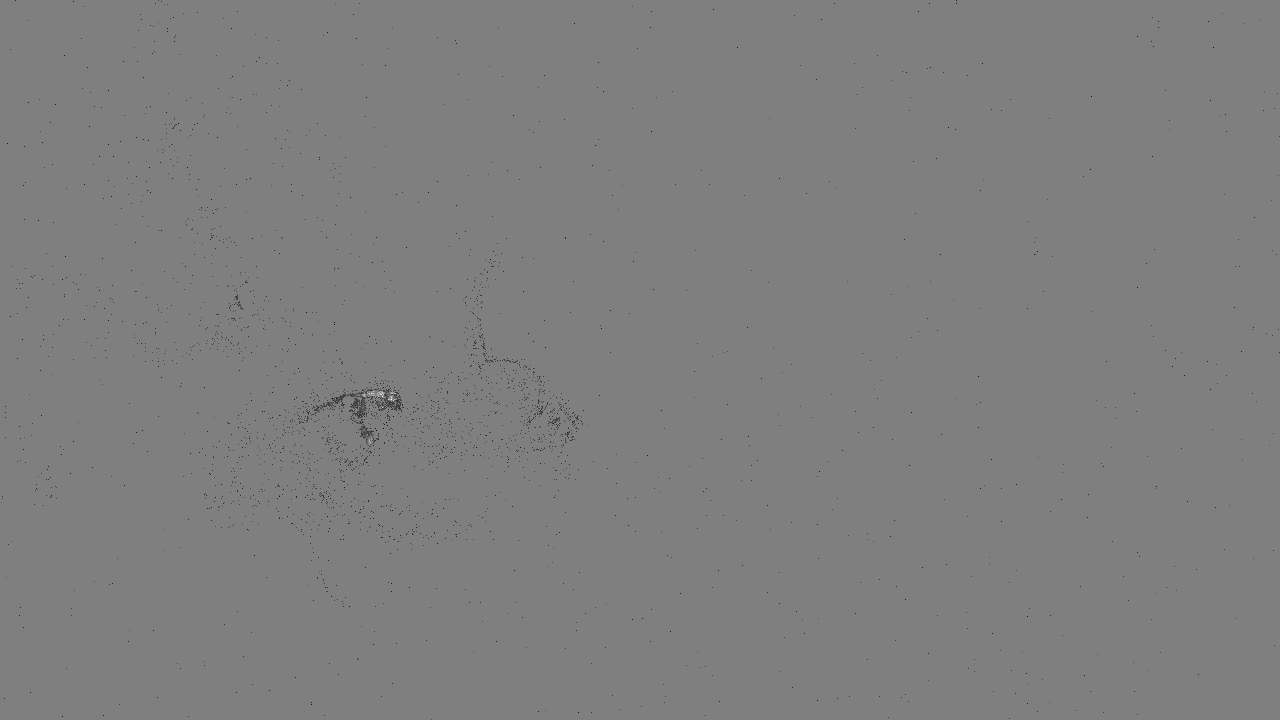} &
        \includegraphics[width=0.45\linewidth, keepaspectratio]{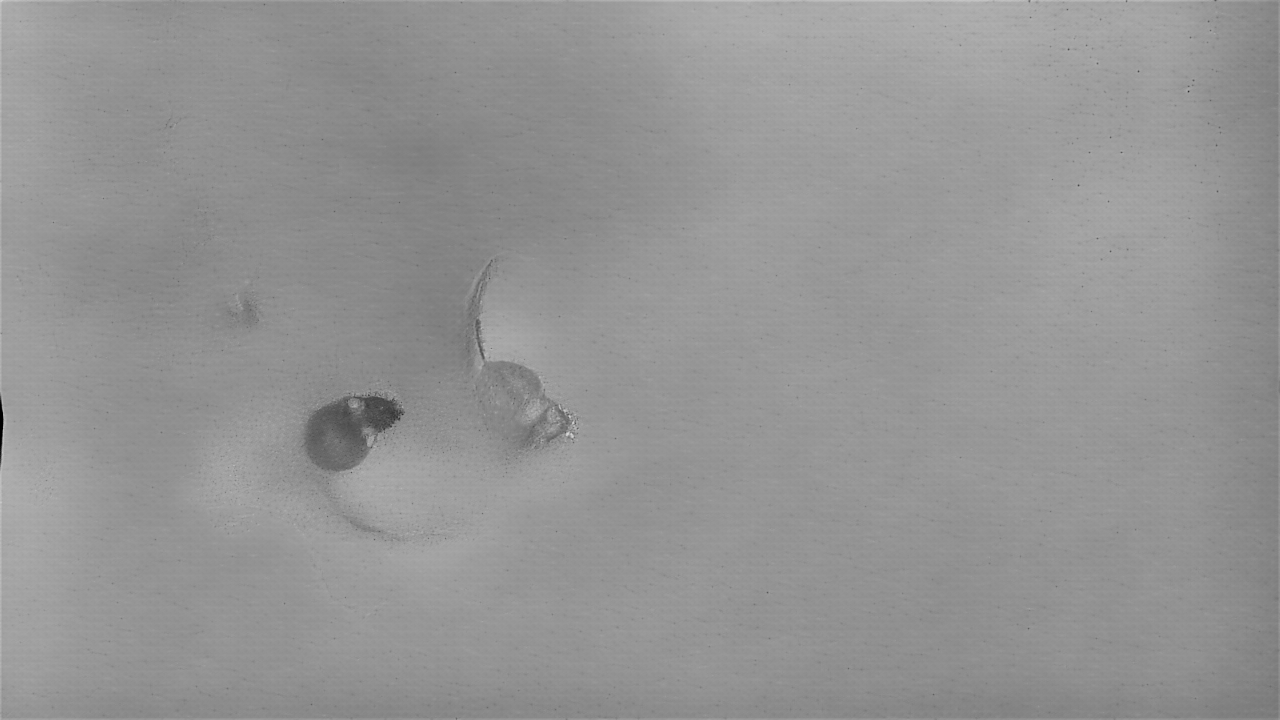} \\
        (c) Voxel &
        (d) Reconstruction
    \end{tabular}
    \caption{(\emph{Team 4}). Illustration of (a) events, (b) GMM clustering result, (c) voxelized events, and (d) final reconstructed frame.}
    \label{fig:frame_comparison:two}
\end{figure}

The detector is a YOLOv8n model initialized with MouseSIS YOLO e2vid pretrained weights and fine-tuned on those reconstructed frames. 
Training runs for 300 epochs with batch size of 32, an initial learning rate of 0.001, a final learning rate of 0.0001, and weight decay is $5 \cdot 10^{-4}$ on two RTX 3090 GPUs.

The team adopts SAM2 as a segmenter. 
YOLO-generated bounding boxes are provided as prompts to SAM2 to generate a binary mask for each frame. 
The resulting masks are produced once and utilized for downstream processing without updating the detector, as shown in \cref{fig:architecture}.

\begin{figure}[tbh]
    \centering
    \includegraphics[width=\linewidth,trim={1.5cm 1.0cm 0.5cm 1.0cm},clip]{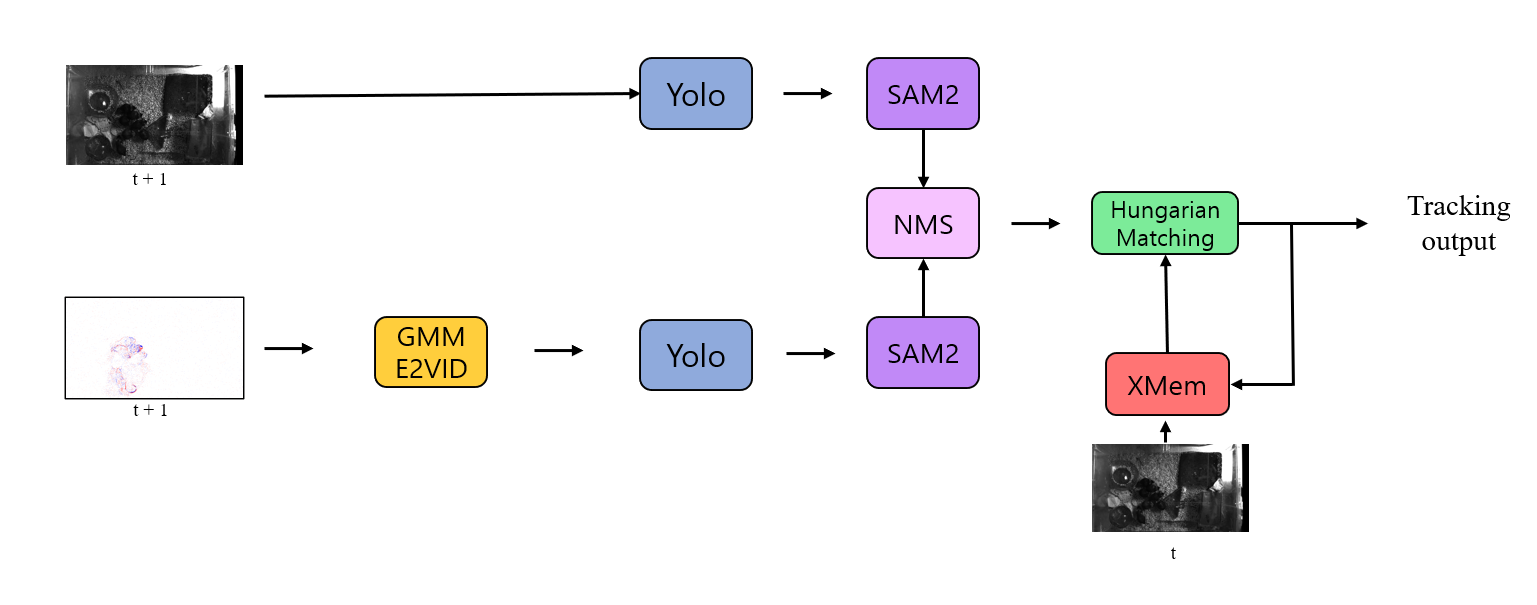}
    \caption{\emph{Team 4}. Architecture overview of the proposed method.}
    \label{fig:architecture}
\end{figure}

\subsubsection{Results}
The method achieves a combined HOTA of 54.01 on the test set. Detailed per‐sequence results are shown in Table~\ref{tab:per_sequence_results}. 

\begin{table}[ht]
    \centering
    \begin{adjustbox}{max width=\linewidth}
    \begin{tabular}{lccccc}
        \toprule
        \textbf{Sequence} & \textbf{MOTA}$\uparrow$   & \textbf{IDF1}$\uparrow$   & \textbf{HOTA}$\uparrow$   & \textbf{DetA}$\uparrow$    & \textbf{AssA}$\uparrow$   \\ 
        \midrule
        10       & 0.754 & 0.647 & 0.536 & 0.619  & 0.470 \\ 
        16       & 0.592 & 0.718 & 0.501 & 0.533  & 0.471 \\ 
        22       & 0.425 & 0.568 & 0.387 & 0.442  & 0.343 \\ 
        26       & 0.242 & 0.438 & 0.404 & 0.415  & 0.408 \\ 
        28       & 0.544 & 0.756 & 0.576 & 0.530  & 0.626 \\ 
        32       & 0.913 & 0.954 & 0.741 & 0.728  & 0.756 \\ 
        Combined & 0.606 & 0.677 & 0.540 & 0.549  & 0.535 \\ 
        \bottomrule
    \end{tabular}
    \end{adjustbox}
    \caption{(\emph{Team 4}). Per‐sequence and combined results for MOTA, IDF1, HOTA, DetA, and AssA.}
    \label{tab:per_sequence_results}
\end{table}

    \subsection{Team 5: vivien}

\subsubsection{Description}
Space-time Instance Segmentation (SIS) is crucial for detailed behavioral analysis, particularly in studies involving laboratory animals like mice.
The MouseSIS dataset \cite{Hamann24mousesis}, used in this challenge, provides rich multi-modal data from grayscale frames and event-based cameras, whose advantages have been extensively surveyed \cite{Gallego20pami}.
The official ModelMixSort baseline \cite{Hamann24mousesis} employs a tracking-by-detection paradigm inspired by methods like SORT \cite{Bewley16icip}, integrating YOLOv8 for object detection, SAM for instance segmentation and XMemSort for tracking.

The team's work was motivated by the opportunity to enhance several aspects of this baseline:
\begin{itemize}
    \item \textbf{Segmentation Model and Mask Quality:} The baseline SAM implementation relies on the \texttt{`transformers'} library and \texttt{`facebook/sam-vit-huge'} \cite{Kirillov23iccv}, selecting a mask from multiple predictions based on IoU scores. The team hypothesized that utilizing a more recent SAM variant, SAM2.1, through the \texttt{`ultralytics.SAM'} implementation with its direct mask output, could offer improved segmentation accuracy and potentially more consistent mask quality.
    \item \textbf{Tracking Persistence:} The XMemSort tracker in the baseline has a \texttt{max\_age} of 1, meaning a track is terminated if it's not matched for more than a single frame. For dynamic mouse movements, this might be too aggressive, leading to premature track termination and increased ID switches. The team aimed to optimize this parameter for better tracking continuity.
\end{itemize} 

\subsubsection{Implementation Details}
The method adheres to the tracking-by-detection framework established by ModelMixSort. The overall pipeline is illustrated in \cref{fig:workflow}.

The key stages are:
\begin{figure}[t!] 
    \centering
    \includegraphics[width=\linewidth]{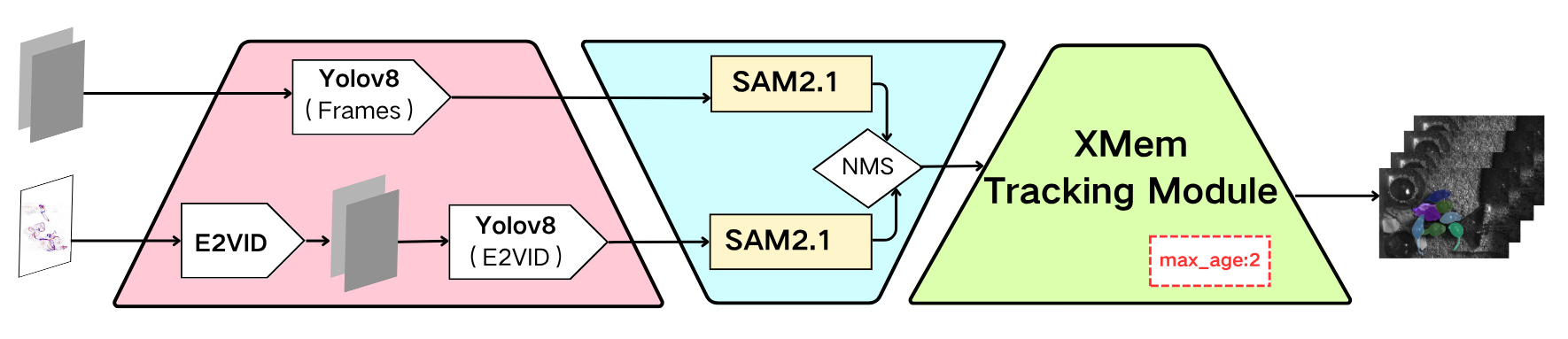} %
    \caption{\emph{Team 5}. Overall pipeline of the method, highlighting modifications to the ModelMixSort baseline.}
    \label{fig:workflow}
\end{figure}
\begin{enumerate}
    \item \textbf{Data Preprocessing:} Event streams are converted into intensity frames using E2VID \cite{Rebecq19pami}.
    \item \textbf{Dual-Path Detection \& Segmentation:}
        \begin{itemize}
            \item YOLOv8 \cite{Jocher23yolov8} detects mice in grayscale frames and E2VID frames, providing bounding box proposals.
            \item SAM2.1 \cite{sam2_1_ultralytics} takes these bounding boxes as prompts to generate precise instance segmentation masks for each detected mouse.
        \end{itemize}
    \item \textbf{Detection Fusion:} Masks from the grayscale and E2VID pathways are merged using Non-Maximum Suppression (NMS).
    \item \textbf{Tracking:} The fused instance masks are fed into the XMemSort tracker (an adaptation of XMem \cite{Cheng22cvpr} with a SORT-like management logic).
    \item \textbf{Output:} The final output consists of per-frame instance segmentation masks with consistent temporal IDs.
\end{enumerate}

\paragraph{Data Preprocessing.}
The primary preprocessing step involves converting the raw event data from the DVS camera into a sequence of intensity frames. This is achieved using the E2VID model \cite{Rebecq19pami}, as provided in the baseline framework. These E2VID-reconstructed frames are then synchronized with the standard grayscale video frames for subsequent processing. No other significant modifications were made to the data preprocessing stage of the baseline.\\

\noindent \textbf{Detection Module (with SAM2.1).}
The detection module is responsible for identifying and segmenting individual mice in each frame from both grayscale and E2VID modalities.

\textbf{Initial Object Detection with YOLOv8.}
Following the baseline, the team employs two instances of the YOLOv8 object detector:
\begin{itemize}
    \item One detector is applied to the standard grayscale frames.
    \item Another detector is applied to the E2VID-reconstructed frames.
\end{itemize}
Both detectors are pre-trained and produce bounding boxes corresponding to potential mouse instances. 
These boxes serve as prompts for the subsequent segmentation stage.

\textbf{Accurate Instance Segmentation with SAM2.1.}
A core modification in the approach is the change in the SAM implementation and model. The baseline utilizes \texttt{`facebook/sam-vit-huge'} \cite{Kirillov23iccv}, selecting the final mask from multiple predictions based on SAM's internally predicted IoU scores.\\
The team replaced this with the \texttt{`ultralytics.SAM'} implementation. The YOLOv8-generated bounding boxes are directly used as prompts for the \texttt{`predict'} method of this SAM2.1 model, which then outputs the final segmentation masks. This change aims to leverage the potentially improved segmentation performance and a more streamlined mask generation process of the \texttt{`ultralytics.SAM'} library with SAM2.1.\\

\noindent \textbf{Tracking Module (with Refined XMem configuration).}
The tracking module associates the fused instance masks across consecutive frames, assigning consistent temporal IDs. 
The team uses the XMemSort tracker, which integrates the XMem video object segmentation model with a SORT-like track management system.

\textbf{Core Tracking Mechanism.}
XMemSort takes the set of unique instance masks from the detection fusion stage as input for each frame. 
It maintains a set of active tracks and attempts to associate new detections with existing tracks based on mask IoU. For matched tracks, XMem updates the mask. For unmatched detections, new tracks are initialized.

\textbf{Parameter Optimization.}
The primary optimization in the tracking module involves the \texttt{max\_age} parameter of XMemSort. 
This parameter defines the maximum number of consecutive frames a track can remain unmatched before it is terminated.
\begin{itemize}
    \item \textbf{Baseline \texttt{max\_age}:} The baseline configuration sets \texttt{tracker:max\_age} to \texttt{1}.
    \item \textbf{Team's \texttt{max\_age}:} The team increased this value to 2 in their configuration.
\end{itemize}
The rationale for increasing \texttt{max\_age} from 1 to 2 is to provide slightly more persistence to tracks. 
With \texttt{max\_age=1}, a track is deleted if it is not matched in the very next frame. 
By changing it to 2, a track can survive one missed frame and potentially be re-associated in the subsequent frame. 
This can be beneficial for handling very brief occlusions or momentary detection failures, thus improving track continuity for the dynamic movements of mice.

The \texttt{min\_hits} parameter (number of consecutive matches required to activate a track) was kept at 3, and the \texttt{iou\_threshold} for matching detections to tracks within XMemSort (configured under \texttt{tracker:iou\_threshold}) was kept at 0.3, consistent with the baseline.\\

\noindent \textbf{Inference Procedure}
\begin{enumerate}
    \item The script loads the configuration from the specified YAML file, which defines data paths, model paths, and key parameters like NMS IoU threshold and tracker settings.
    \item For each sequence in the specified split :
        \begin{itemize}
            \item Grayscale frames are loaded from HDF5 files.
            \item E2VID frames are pre-loaded from their respective directories.
            \item The \texttt{SamYoloDetector} (utilizing YOLOv8 and the team's SAM2.1 setup) is instantiated separately for grayscale and E2VID data.
            \item The XMemSort tracker is initialized with the team's optimized parameters (\texttt{max\_age: 2}, \texttt{min\_hits: 3}, \texttt{iou\_threshold: 0.3}).
            \item Per frame, detections are obtained from both grayscale and E2VID paths using the respective detectors.
            \item These detections are fused using NMS.
            \item The fused masks are passed to the XMemSort tracker, which updates track states and assigns IDs.
            \item Results (segmentation masks and track IDs per frame) are formatted into the COCO-like JSON structure.
        \end{itemize}
    \item Finally, results from all processed sequences are aggregated into a single \texttt{final\_results.json} file for submission and evaluation.
    \item Visualization of predictions per frame was enabled during processing for debugging and qualitative assessment.
\end{enumerate}

\subsubsection{Results}
\noindent \textbf{Experimental Setup}
\begin{itemize}
    \item \textbf{Dataset:} The experiments were conducted on the \mbox{MouseSIS} Challenge Dataset \cite{Hamann24mousesis}. 
    \item \textbf{Evaluation Metrics:} Performance was evaluated using standard multi-object tracking metrics, including HOTA~\cite{Luiten21ijcv}, CLEAR MOT \cite{Luiten21ijcv}, and IDF1 \cite{Ristani16eccv}.
    \item \textbf{Software Environment:} The implementation is based on Python. 
    Key libraries include PyTorch for deep learning models, Ultralytics (for YOLOv8 and SAM2.1), OpenCV for image processing, NumPy for numerical operations, h5py for data loading, and PyYAML for configuration management. 
    All experiments were run on a Linux operating system.
    \item \textbf{Hardware:} Experiments were conducted on a system equipped with a 14-core Intel(R) Xeon(R) Gold 6330 CPU @ 2.00GHz, 90 GB of RAM, and a single NVIDIA RTX 3090 GPU with 24GB VRAM. 
\end{itemize}

\paragraph{Main Challenge Performance.}
The team's performance on the official test set of the SIS Challenge is reported in Tab.~1. %

\paragraph{Ablation Studies and Parameter Analysis.}
All ablation studies were conducted on the combined MouseSIS validation sequences (03, 04, 12, 25).

\textbf{Impact of SAM2.1 vs.~Original SAM}:
To isolate the effect of upgrading the segmentation model, the team compared the original SAM with their modification using SAM2.1 (\texttt{ultralytics.SAM}) while keeping other key parameters consistent with the original baseline configuration. 
The results are shown in \cref{tab:sam_ablation}.
The introduction of SAM2.1 led to substantial improvements across all major metrics, with HOTA increasing by 24.44\% and IDF1 by 28.00\%. This highlights the significant benefit of the newer segmentation model and the team's chosen implementation for this task.
\begin{table}[ht]
    \centering
    \begin{adjustbox}{max width=\linewidth}
        \begin{tabular}{lccc}
            \toprule
            \textbf{Configuration} & \textbf{HOTA}$\uparrow$ & \textbf{MOTA}$\uparrow$ & \textbf{IDF1}$\uparrow$ \\
            \midrule
            Baseline (SAM) & 0.45 & 0.62 & 0.50 \\
            Team's (SAM2.1)  & \textbf{0.56} & \textbf{0.74} & \textbf{0.64} \\
            \midrule
            Improvement (\%) & +24.44 & +19.35 & +28.00 \\
            \bottomrule
        \end{tabular}
    \end{adjustbox}
    \caption{(\emph{Team 5}). 
    Ablation Study: Original SAM vs.~SAM2.1 (Validation Set, Combined).}
    \label{tab:sam_ablation}
\end{table}

\textbf{Sensitivity to XMemSort's \texttt{max\_age} Parameter}:
The team investigated the impact of the XMemSort's \texttt{max\_age} parameter. 
These experiments were conducted using SAM2.1. 
The baseline configuration for \texttt{max\_age} was 1. 
The team tested \texttt{max\_age} values of 2 and 3.

\begin{table}[ht]
    \centering
    \begin{adjustbox}{max width=\linewidth}
    \begin{tabular}{lccccc}
        \toprule
        \texttt{max\_age} & \textbf{HOTA}$\uparrow$ & \textbf{AssA}$\uparrow$ & \textbf{IDF1}$\uparrow$ & \textbf{MOTA}$\uparrow$ & \textbf{IDSW}$\downarrow$ \\
        \midrule
        1  & 0.557 & 0.512 & 0.638 & 0.740 & 96 \\
        2  & 0.560 & 0.516 & 0.657 & 0.742 & 80 \\
        3  & \textbf{0.583} & \textbf{0.557} & \textbf{0.693} & \textbf{0.744} & \textbf{62} \\
        \bottomrule
    \end{tabular}
    \end{adjustbox}
    \caption{(\emph{Team 5}). 
    Sensitivity Analysis: XMemSort \texttt{max\_age} (Validation Set, Combined).}
    \label{tab:maxage_sensitivity}
\end{table}

Increasing \texttt{max\_age} from 1 to 2 showed a slight improvement in HOTA and IDF1, and a reduction in Identity Switch (IDSW). 
Further increasing \texttt{max\_age} to 3 yielded the best validation performance. 
This suggests that allowing tracks to persist for a slightly longer duration (2--3 frames) when unmatched is beneficial for this dataset, likely by better handling brief occlusions or misdetections without losing track identity. 
The team's final submitted model used \texttt{max\_age=2} as a balance, though \texttt{max\_age=3} showed superior validation results.

\else
    \section{Challenge Teams and Methods}
    \label{sec:team_methods}

\fi

\section*{Acknowledgments}
Funded by the Deutsche Forschungsgemeinschaft (DFG, German Research Foundation) under Germany’s Excellence Strategy – EXC 2002/1 ``Science of Intelligence'' – project number 390523135.

{
    \small
    \bibliographystyle{ieeenat_fullname}

}

\end{document}